\documentclass[conference, a4paper]{IEEEtran}
\usepackage{graphicx}

\usepackage{CJKutf8}

\usepackage{multirow}

\usepackage{amsmath}
\usepackage{amsfonts}

\usepackage{comment}
\usepackage{booktabs}
\usepackage{dirtytalk}
\usepackage{cite}

\usepackage{threeparttable} 
\usepackage{xcolor} 

\makeatletter
\let\MYcaption\@makecaption
\makeatother

\usepackage[font=footnotesize]{subcaption}

\makeatletter
\let\@makecaption\MYcaption
\makeatother

\usepackage{float}

\ifCLASSINFOpdf
\else
\fi

\begin{document}

\title{Hierarchical Object Detection and Recognition Framework
for Practical Plant Disease Diagnosis
}

\makeatletter
\newcommand{\linebreakand}{%
  \end{@IEEEauthorhalign}
  \hfill\mbox{}\par
  \mbox{}\hfill\begin{@IEEEauthorhalign}
}
\makeatother

\author{
    \IEEEauthorblockN{
        Kohei Iwano\textsuperscript{\rm 1}, 
        Shogo Shibuya\textsuperscript{\rm 1}, 
        Satoshi Kagiwada\textsuperscript{\rm 2}, 
        Hitoshi Iyatomi\textsuperscript{\rm 1}
        }
    \IEEEauthorblockA{
        \textsuperscript{\rm 1}Applied Informatics, Graduate School of Science and Engineering, Hosei University, Tokyo, Japan\\
        \textsuperscript{\rm 2}Clinical Plant Science, Faculty of Bioscience and Applied Chemistry, Hosei University, Tokyo, Japan\\
        kohei.iwano.0528@gmail.com, shogo.shibuya.5u@gmail.com,  kagiwada@hosei.ac.jp, iyatomi@hosei.ac.jp
        }
}

\maketitle

\begin{abstract}
    Recently, object detection methods (OD; e.g., YOLO-based models) have been widely utilized in plant disease diagnosis. 
These methods demonstrate robustness to distance variations and excel at detecting small lesions compared to classification methods (CL; e.g., CNN models).
However, there are issues such as low diagnostic performance for hard-to-detect diseases and high labeling costs.
Additionally, since healthy cases cannot be explicitly trained, there is a risk of false positives.
We propose the Hierarchical object detection and recognition framework (HODRF), a sophisticated and highly integrated two-stage system that combines the strengths of both OD and CL for plant disease diagnosis.
In the first stage, HODRF uses OD to identify regions of interest (ROIs) without specifying the disease.
In the second stage, CL diagnoses diseases surrounding the ROIs.
HODRF offers several advantages: 
(1) Since OD detects only one type of ROI, HODRF can detect diseases with limited training images by leveraging its ability to identify other lesions.
(2) While OD over-detects healthy cases, HODRF significantly reduces these errors by using CL in the second stage.
(3) CL's accuracy improves in HODRF as it identifies diagnostic targets given as ROIs, making it less vulnerable to size changes.
(4) HODRF benefits from CL's lower annotation costs, allowing it to learn from a larger number of images.
We implemented HODRF using YOLOv7 for OD and EfficientNetV2 for CL and evaluated its performance on a large-scale dataset (4 crops, 20 diseased and healthy classes, 281K images).
HODRF outperformed YOLOv7 alone by 5.8 to 21.5 points on healthy data and 0.6 to 7.5 points on macro F1 scores, and it improved macro F1 by 1.1 to 7.2 points over EfficientNetV2.

\end{abstract}


\IEEEpeerreviewmaketitle

\section{Introduction}
    In recent years, automatic diagnostic methods using images have been developed to reduce the time and cost required to diagnose plant diseases and insect pests, and to detect them at an early stage. Particularly, classification-based diagnostic methods (CL),  represented by convolutional neural networks (CNNs), have many advantages, including ease of training, and many good results have been reported~\cite{hughes2015open,mohanty2016using,wang2017automatic,toda2019convolutional,atila2021plant,Kawasaki2015,ramcharan2017deep,fujita2018practical,elfatimi2022beans,narayanan2022banana,shibuya2021validation,saikawa2019aop,kanno2021,boulent2019convolutional,xu2023embracing}.
In the early days, a large open dataset, the PlantVillage Dataset~\cite{hughes2015open}, was often used. This early dataset consisted of images taken from experimental (laboratory) environments, and while models using it showed excellent accuracy within the dataset, classification accuracy was reportedly significantly degraded when the models were compared to images taken in real environments~\cite{mohanty2016using}.
Since then, various real-environment datasets have been made publicly available, and more studies have been conducted based on images collected from real environments~\cite{Kawasaki2015,ramcharan2017deep,fujita2018practical,elfatimi2022beans,narayanan2022banana,shibuya2021validation,saikawa2019aop,kanno2021,boulent2019convolutional,xu2023embracing}.

\begin{figure*}[t]
    \begin{center}
    \includegraphics[width=150mm]{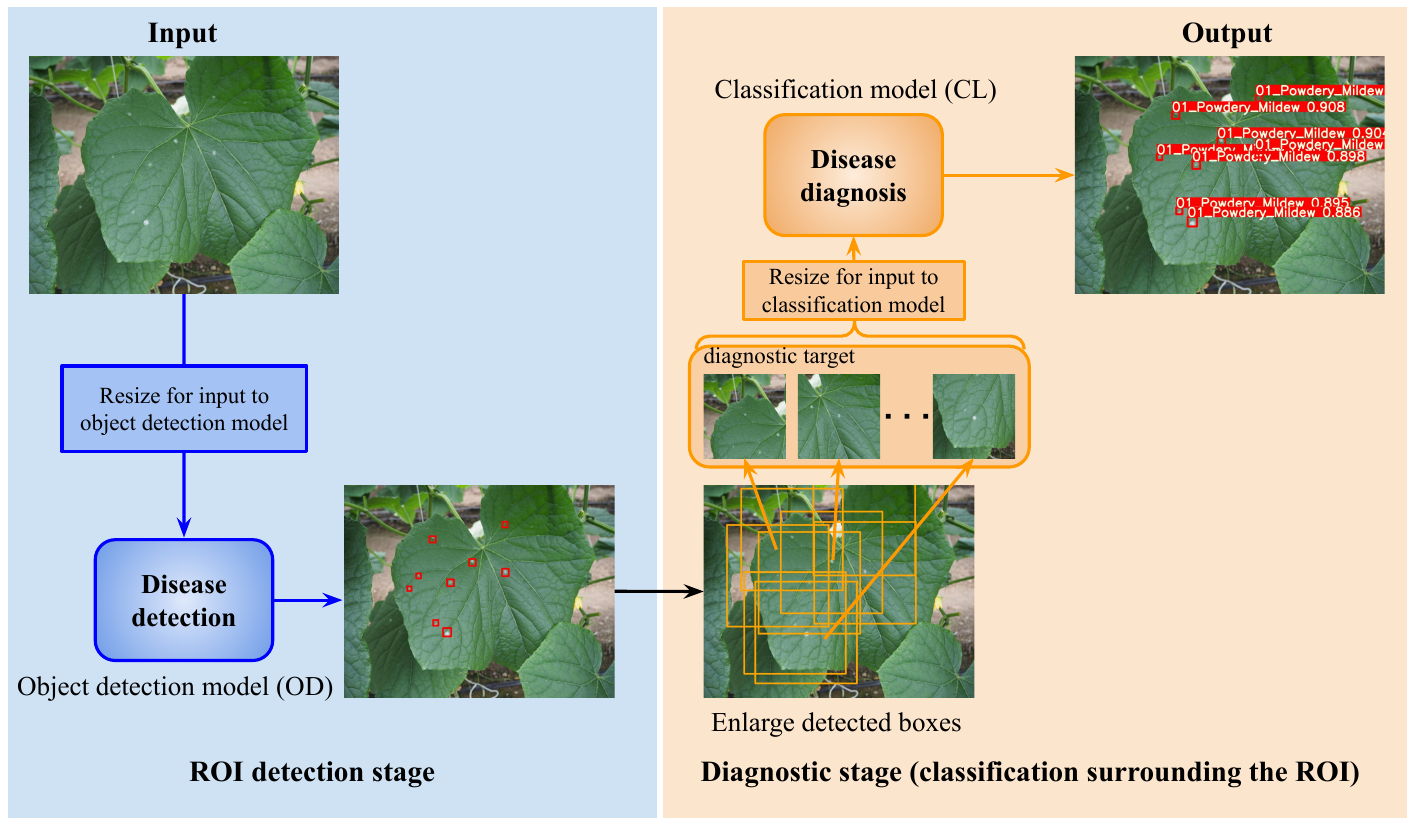}
    \caption{Diagram of the HODRF}
    \label{fig:HODRF}
    \end{center}
\end{figure*}
Many studies using practical images for training have reported excellent diagnostic accuracy. However, concerns have been raised in recent years that the collected images are not properly divided into training and evaluation images, resulting in ``raising'' numbers beyond the intrinsic performance being reported~\cite{shibuya2021validation,saikawa2019aop,kanno2021,boulent2019convolutional,xu2023embracing}.
In plant disease diagnosis, the disease symptom areas that provide diagnostic clues are often extremely small, and because of the diversity of disease symptoms, the discriminator is susceptible to disturbances other than disease symptoms.
For example, subjects photographed in the same field often have high image similarities due to a similar composition, background, and disease symptom appearance and extent, including distance and angle between the camera and the subject.
Therefore, storing the collected labeled data in a single data set and dividing it into train, validation, and test sets in arbitrary proportions, as is done in many studies, or submitting it to cross-validation, etc., will increase its numerical diagnostic ability, but its actual ability will deviate.
In 2021, Shibuya et al.~\cite{shibuya2021validation} analyzed an unprecedentedly large dataset of 221K images of four crops from 27 strictly controlled fields, using the most advanced CNN at the time, and they showed that overfitting on the training data and insufficient generalizability to disease could not be achieved.
We have shown that this is mainly a representative case of CL, primarily because CNNs, which are representative examples of CL, are not robust to variations in the distance to the subject, and the training data are not fully trained on a wide variety of images to cover the unknown variety of test data.

Conversely, methods that apply deep learning-based object detection (OD) techniques, such as single shot multibox detector (SSD)~\cite{ssd}, and YOLOv7~\cite{wang2023yolov7} are robust to differences in distance to the subject and have a strong performance in detecting small lesions. In addition to diagnostic accuracy, high value-added results, such as readable results, have been reported~\cite{fuentes2017robust,chen2022plant,suwa2019}.
In addition, because OD can focus on local regions, it is expected to suppress the effects of image background~\cite{saikawa2019aop,wayama2023investigation} and subject composition, one factor in overfitting.
However, the cost of annotating the OD teacher data is extremely high, resulting in a limited set of images that can be used for training, in turn leading to an inherently poor diagnostic capability with unknown data~\cite{suwa2019}.
In addition, the greatest challenge with OD is its inability to utilize data explicitly with no detectable parts (i.e., healthy data) for training, resulting in low discriminative power for healthy cases (often over-detection of disease)~\cite{cap2023towards,okamoto2021gastric}. It also tends toward poor diagnosis of targets needed to capture the signs of overall disease~\cite{wayama2023investigation}. This further leads to difficulty in detecting viral diseases, for which early detection is particularly important.
Suwa et al.~\cite{suwa2019} showed that in a diseased cucumber leaf detection task in a wide-area field, a system based on SSD, the most advanced OD at the time, directly caused the aforementioned overfitting and was extremely problematic in its capacity for diagnosing unknown data.
They then proposed a two-stage diagnostic system in which the leaf area that serves as the ROI is detected by an OD model, followed by diagnosis in an independent stage, demonstrating certain results.
Their two-stage diagnostic model was an effective framework, but its evaluation was limited to only two classifications of a single crop (cucumber): diseased and healthy.
Thus, the two-step diagnostic model is a highly practical framework for realizing automated diagnostic applications, but a systematic and practical system construction and an evaluation using large-scale data have not been conducted.
\\\indent 
In this paper, we propose a practical hierarchical object detection and recognition framework (HODRF) for the automatic diagnosis of plant diseases consisting of two stages that effectively combine OD and CL.
The first stage of HODRF uses OD to detect the ROI, in this case the diseased area of the leaf, and the second stage uses a disease discriminator with CL that adopts the area around the detected diseased area as the input for diagnosis.
Because convolutional neural network (CNN), which is widely used as a general CL, has a low annotation cost and can be trained at a low cost on a larger number of images, including of healthy leaves, for which OD models cannot be explicitly trained, this two-stage system is expected to be effective in compensating for the weaknesses of both. 
In this experiment, YOLOv7~\cite{wang2023yolov7} and EfficientNetV2~\cite{tan2021efficientnetv2}, currently the most advanced OD and CL, are used to configure the proposed HODRF.
The model was validated using a large image dataset of leaves of four crops (strawberry, eggplant, tomato, and cucumber) with high-quality disease labels (a total of 21 classes including healthy; 241,214 training images; and 40,464 images for evaluation) collected from 27 fields in 24 prefectures in Japan.

\section{Hierarchical object detection and recognition framework (HODRF)}
    %
%
%

%
%
%
%
%

\begin{figure}[t]
    \centering
    \begin{minipage}{0.375\linewidth}
        \includegraphics[width=\linewidth]{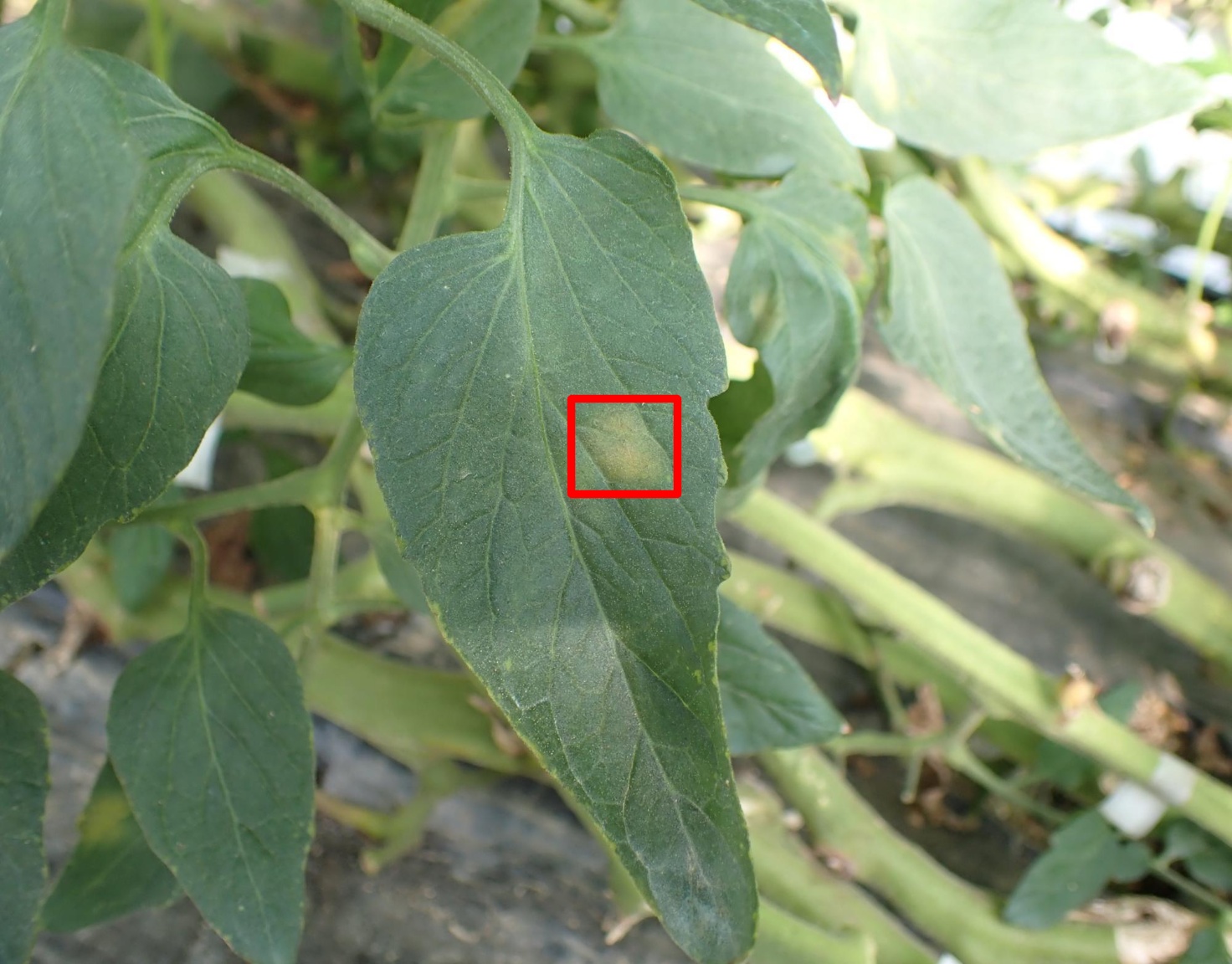}
        \subcaption{ROI detected in the ROI \\ detection stage}
    \end{minipage}
    \begin{minipage}{0.295\linewidth}
        \includegraphics[width=\linewidth]{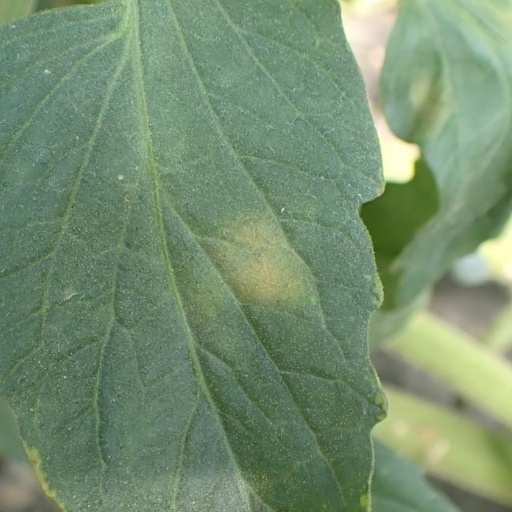}
        \subcaption{Detected diagnosis \\ target based on ROI}
    \end{minipage}
    \begin{minipage}{0.295\linewidth}
        \includegraphics[width=\linewidth]{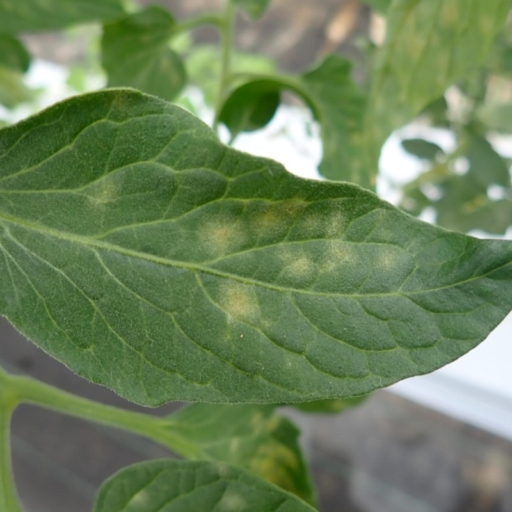}
        \subcaption{Example image \\ trained for the CL}
    \end{minipage}
   \caption{Example of a detected ROI, its enlarged ($l \times l$ pixels) diagnostic target, and an image used to train the CL (4\_CLM; tomato).}
   \label{fig:examples}
\end{figure}
%

In this paper, we propose an HODRF, a sophisticated and highly integrated two-stage diagnostic framework that effectively combines OD and CL techniques.
Fig.~\ref{fig:HODRF} shows a diagram of the HODRF.
The first ROI detection stage uses an OD to detect the ROIs containing the lesion, and the second diagnostic stage uses a CL to diagnose the area containing the ROI. In fact, the choice of model between OD and CL is arbitrary in HODRF.
\subsection{ROI detection stage}
In the ROI detection stage, from the $x^{1st} \times y^{1st}$ pixels input image, the OD detects ROIs, each of which is passed to a later diagnostic stage. Note that the ROIs detected by the OD do not distinguish specific diseases, but rather target all diseases.
Viral diseases often cause disease symptoms to appear across the entire leaf, in which case the entire leaf is the ROI.
If no ROI is detected, it is treated as ``healthy,'' and the process is not passed to CL and is terminated. Because the absence of disease can also be determined at the next stage, only obvious cases are considered healthy at this stage.
\subsection{Diagnostic stage}
In the diagnostic stage, for each ROI detected in the previous stage, the center coordinates of the ROI box are calculated, and a square of $l \times l$ pixels centered on the coordinates is cut out as the diagnostic target area. Each diagnostic target area is resized to $x^{2nd} \times y^{2nd}$ pixels using bi-linear interpolation, and each resized region is sequentially inputted into CL, producing a diagnostic result. Our experiments do not deal with cases of multiple infections, so when multiple ROIs are present, the majority CL output is used as the diagnostic result. When dealing with multiple infections, it is possible to list all diseases that exceed the thresholds set for the CL outputs, but this is beyond the scope of this paper.
Fig.~\ref{fig:examples} shows an example of the ROI determined with the OD (a), its extended diagnosis target (b), and an example image trained by the CL (c) to diagnose targets like (b).

\section{Experiments}
    \subsection{Dataset}
In this paper, we used 281,678 images in total of four crops. They were independently grown, inoculated with disease, photographed, and professionally labeled under strict disease control at 27 agricultural experiment stations in 24 prefectures in Japan from 2016 to 2020. The images are generally composed of a single leaf near the center, but there are also many images with multiple leaves at a variety of distances from the subject. Our experimental setup differs from many previous studies in that the training and test images are taken at different locations and rigorously evaluated against different and completely unknown data. Table~\ref{table1} outlines the training data (including the number of bounding boxes with annotations used in the OD training) and test data used in the study. Diseases treated were: 1: Powdery Mildew (PM), 2: Gray Mold (GM), 3: Anthracnose (AN), 4: Cercospora Leaf Mold (CLM), 5: Leaf Mold (LM), 6: Late Blight (LB), 7: Downy Mildew (DM), 8: Corynespora Leaf Spot (CLS), 9: Corynespora Target Spot (CTS), 10: Leaf Spot (LS), 11: Fusarium Wilt (FW), 12: Gummy Stem Blight (GSB), 13: Verticillium Wilt (VW), 14: Bacterial Wilt (BW), 15: Bacterial Spot (BS), 16: Bacterial Canker (BC), 17: Cucurbit Chlorotic Yellows Virus (CCYV), 18: Mosaic Diseases (MD), 19: Melon Yellow Spot Virus (MYSV), 20: Yellow Leaf Curl (YLC). This produced a total 21 classes, comprising 20 diseased (17-20 are viral diseases) and 0: Healthy (HE).
%
%
\begin{table*}[t]
\caption{Details of the dataset}
    \begin{minipage}[]{0.50\textwidth}
    \vspace{-27pt}
    \begin{center}
        \begin{tabular*}{7.9cm}{@{\extracolsep{\fill}}lrrrrr@{}}
            \toprule
            \multicolumn{5}{c}{(a) Strawberry (4 class)} \\ \midrule
            \multirow{2}{*}{disease class} & \multicolumn{2}{c}{YOLOv7}                                  & EfficientNetV2   & \multicolumn{1}{c}{\multirow{2}{*}{Test}} \\ \cmidrule(lr){2-3}\cmidrule(lr){4-4}
                                        & \multicolumn{1}{c}{Train} & \multicolumn{1}{c}{\# of boxes} & Train & \multicolumn{1}{c}{}                      \\ \midrule
            0\_HE   & N/A    & N/A     & 10,472  & 578   \\
            1\_PM   & 735    & 5,245   & 1,952   & 893   \\
            3\_AN   & 1,459  & 11,619  & 3,701   & 609   \\
            11\_FW  & 538    & 800     & 2,608   & 227   \\ \midrule
            \textbf{Total}   & \textbf{2,732}  & \textbf{17,664}  & \textbf{18,733}  & \textbf{2,307} \\ \bottomrule
            \end{tabular*}
    \end{center}
    \end{minipage} 
  \hfill
    \begin{minipage}[]{0.5\textwidth}
    \begin{center}
        \begin{tabular*}{7.9cm}{@{\extracolsep{\fill}}lrrrrr@{}}
            \toprule
            \multicolumn{5}{c}{(b) Eggplant (7 class)} \\ \midrule
            \multicolumn{1}{c}{\multirow{2}{*}{disease class}} & \multicolumn{2}{c}{YOLOv7}                  & EfficientNetV2             & \multicolumn{1}{c}{\multirow{2}{*}{Test}} \\ \cmidrule(lr){2-3}\cmidrule(lr){4-4}
            \multicolumn{1}{c}{}                            & \multicolumn{1}{c}{Train} & \# of boxes     & Train           & \multicolumn{1}{c}{}                      \\ \midrule
            0\_HE   & N/A    & N/A    & 12,335 & 1,122 \\
            1\_PM   & 1,706  & 4,468  & 7,936  & 938   \\
            2\_GM   & 513    & 782    & 1,024  & 166   \\
            5\_LM   & 1,191  & 14,801 & 3,188  & 732   \\
            10\_LS  & 2,399  & 1,980  & 5,509  & 118   \\
            13\_VW  & 1,248  & 1,396  & 3,167  & 353   \\
            14\_BW  & 1,343  & 3,175  & 3,415  & 462   \\ \midrule
            \textbf{Total}                                  & \textbf{8,400}            & \textbf{26,602} & \textbf{36,574} & \textbf{3,891}                            \\ \bottomrule
        \end{tabular*}
    \end{center}
    \end{minipage} 
    \hfill
    \begin{minipage}[c]{0.499\textwidth}
    \begin{center}
    \vspace{3.0mm}
        \begin{tabular*}{7.9cm}{@{\extracolsep{\fill}}lrrrrr@{}}
            \toprule
            \multicolumn{5}{c}{(c) Tomato (10 class)} \\ \midrule
            \multicolumn{1}{r}{\multirow{2}{*}{disease class}} & \multicolumn{2}{c}{YOLOv7}        & EfficientNetV2             & \multicolumn{1}{c}{\multirow{2}{*}{Test}} \\ \cmidrule(lr){2-3}\cmidrule(lr){4-4}
            \multicolumn{1}{r}{}                            & Train           & \# of boxes     & Train           & \multicolumn{1}{c}{}                      \\ \midrule
            0\_HE    & N/A   & N/A    & 8,120 & 2,994 \\
            1\_PM    & 1,778 & 21,748 & 4,490 & 4,250 \\
            2\_GM    & 1,075 & 1,443  & 9,327 & 571   \\
            4\_CLM   & 2,238 & 15,538 & 4,078 & 1,809 \\
            5\_LM    & 2,048 & 15,263 & 2,761 & 151   \\
            6\_LB    & 1,457 & 2,480  & 2,049 & 808   \\
            9\_CTS   & 1,060 & 6,353  & 1,732 & 1,350 \\
            14\_BW   & 1,334 & 3,405  & 2,259 & 412   \\
            16\_BC   & 1,538 & 3,102  & 4,369 & 128   \\
            20\_YLC  & 1,390 & 5,991  & 4,513 & 1,746 \\ \midrule
            \textbf{Total}                                  & \textbf{13,918} & \textbf{75,323} & \textbf{43,698} & \textbf{14,219}                           \\ \bottomrule
        \end{tabular*}
            
    \end{center}
    \end{minipage}
    \hfill
    \begin{minipage}[c]{0.499\textwidth}
    \begin{center}
    \vspace{5.0mm}
        \begin{tabular*}{7.85cm}{@{\extracolsep{\fill}}lrrrrr@{}}
            \toprule
            \multicolumn{5}{c}{(d) Cucumber (10 class)} \\ \midrule
            \multicolumn{1}{c}{\multirow{2}{*}{disease class}} & \multicolumn{2}{c}{YOLOv7}         & EfficientNetV2             & \multicolumn{1}{c}{\multirow{2}{*}{Test}} \\ \cmidrule(lr){2-3}\cmidrule(lr){4-4}
            \multicolumn{1}{c}{}                            & Train           & \# of boxes      & Train           & \multicolumn{1}{c}{}                      \\ \midrule
            0\_HE    & N/A   & N/A     & 16,023 & 5,576 \\
            1\_PM    & 3,494 & 67,200  & 7,653  & 1,898 \\
            3\_AN    & 3,038 & 39,902  & 3,038  & 1,125 \\
            7\_DM    & 2,671 & 36,808  & 6,953  & 2,578 \\
            8\_CLS   & 3,592 & 110,896 & 7,565  & 1,813 \\
            12\_GSB  & 1,051 & 3,564   & 1,483  & 957   \\
            15\_BS   & 1,045 & 17,181  & 3,719  & 3,291 \\
            17\_CCYV & 3,602 & 4,420   & 5,969  & 179   \\
            18\_MD   & 2,190 & 6,398   & 26,861 & 1,626 \\
            19\_MYSV & 2,292 & 2,450   & 17,233 & 1,004 \\ \midrule
            \textbf{Total}                                  & \textbf{20,662} & \textbf{288,819} & \textbf{96,497} & \textbf{20,047}                           \\ \bottomrule
        \end{tabular*}
    \end{center}
    \vspace{0.3mm}
    \end{minipage}
    \label{table1}
    \leftline{
    \hspace{0.5cm}
    All test data were collected in a different field than the training data field.}
\end{table*}
\subsection{HODRF Implementation}
This paper uses YOLOv7-w6~\cite{wang2023yolov7} (YOLOv7) and EfficientNetV2-s~\cite{tan2021efficientnetv2} (EfficientNetV2) for the OD and CL, respectively, as they are currently reported to have the most advanced performances. YOLOv7 was built by fine-tuning a model pre-trained on the COCO dataset~\cite{lin2014microsoft}. Because each image has a different size and aspect ratio, in this experiment, the center of each image was cropped to fit the short edge and resized to $1,472 \times 1,472$ pixels. For the training images, the following augmentation processes were performed: rotation in 90-degree increments, horizontal and vertical flipping, Median blur (image smoothing by median filter), CLAHE~\cite{109340} (histogram equalization), fog simulation, and Cutout~\cite{devries2017improved} (dropout of a rectangular area). As in the original YOLOv7~\cite{wang2023yolov7}, Momentum SGD~\cite{qian1999momentum} was used as the optimization method for training, with a learning rate of $1×10^{-2}$, momentum term of 0.937, and weight decay of $5×10^{-4}$.
EfficientNetV2 was built by fine-tuning the pre-trained models on the ImageNet dataset~\cite{deng2009imagenet}. For augmentation during training, horizontal and vertical flipping and random rotation were applied, and the image size was $512 \times 512$ pixels. Adam~\cite{kingma2014adam} was used for optimization, with a learning rate of $1 \times 10^{-5}$, and a weight decay of $1 \times 10^{-4}$.
In this experiment, $x^{1st} = 1,472$, $ y^{1st} = 1,472$, $x^{2nd} = 512$, $y^{2nd} = 512$, and $l = 512$ were set as parameters.

\subsection{Evaluation basis and comparison method}
In the comparison method for the proposed HODRF, stand-alone EfficientNetV2 and YOLOv7 were used, and for fairness, the parameters and number of training images are the same as for the HODRF. However, only YOLOv7 cannot train healthy data, so they were excluded. The evaluation indices used were the healthy F1-score, the average disease-only F1-score, the macro F1-score, and the accuracy.

\section{Results}
\begin{table*}[t]
\begin{center}
\vspace{-3.7mm}
\caption{Comparison of disease diagnostic performance for the test dataset (\%)}
\begin{tabular*}{17cm}{@{\extracolsep{\fill}}llrrrr@{}}
\toprule
\multicolumn{1}{c}{Target} &
  \multicolumn{1}{c}{Model} &
  \multicolumn{1}{c}{F1-score (Healthy)} &
  \multicolumn{1}{c}{Avg. F1-score (Diseases)} &
  \multicolumn{1}{c}{Macro F1-score} &
  \multicolumn{1}{c}{Micro Accuracy} \\ \midrule
                                     & EfficientNetV2   & \textbf{77.6}                        & \textbf{87.2} & \textbf{84.8} & \textbf{86.9} \\
                                     & YOLOv7           & 66.3                                 & 84.2          & 79.7          & 84.0          \\
\multirow{-3}{*}{Strawberry (4 class)} &
  HODRF (proposed) &
  {\color[HTML]{FE0000} \textbf{80.4}} &
  {\color[HTML]{FE0000} \textbf{89.4}} &
  {\color[HTML]{FE0000} \textbf{87.2}} &
  {\color[HTML]{FE0000} \textbf{88.8}} \\ \midrule
                                     & EfficientNetV2   & {\color[HTML]{FE0000} \textbf{85.6}} & \textbf{75.0} & \textbf{76.5} & {\color[HTML]{FE0000}\textbf{83.1}} \\
                                     & YOLOv7           & 57.4                                 & 73.6          & 71.2          & 70.8          \\
\multirow{-3}{*}{Eggplant (7 class)} & HODRF (proposed) & \textbf{81.1}                        & {\color[HTML]{FE0000} \textbf{77.0}} & {\color[HTML]{FE0000} \textbf{77.6}} &  \textbf{81.4} \\ \midrule
                                     & EfficientNetV2   & {\color[HTML]{FE0000} \textbf{89.3}} & 75.2          & 76.6          & 87.5          \\
                                     & YOLOv7           & 75.5                                 & \textbf{78.3} & \textbf{78.0} & \textbf{87.7} \\
\multirow{-3}{*}{Tomato (10 class)} &
  HODRF (proposed) &
  \textbf{84.7} &
  {\color[HTML]{FE0000} \textbf{79.0}} &
  {\color[HTML]{FE0000} \textbf{79.5}} &
  {\color[HTML]{FE0000} \textbf{87.8}} \\ \midrule
                                     & EfficientNetV2   & \textbf{75.7}                        & 53.9          & 56.1          & 60.8          \\
                                     & YOLOv7           & 63.5                                 & \textbf{62.6} & \textbf{62.7} & \textbf{61.8} \\
\multirow{-3}{*}{Cucumber (10 class)} &
  HODRF (proposed) &
  {\color[HTML]{FE0000} \textbf{76.4}} &
  {\color[HTML]{FE0000} \textbf{63.0}} &
  {\color[HTML]{FE0000} \textbf{63.3}} &
  {\color[HTML]{FE0000} \textbf{67.8}} \\ \bottomrule
\end{tabular*}
\label{table2}
\end{center}
    \leftline{
    \hspace{0.5cm}
    Bold figures in \textbf{\color{red}
    red} and \textbf{black} indicate the best and second-best performances, respectively.}

\end{table*}
%
%
%
\begin{figure*}[t]
    \begin{center}
    \includegraphics[width=125mm]{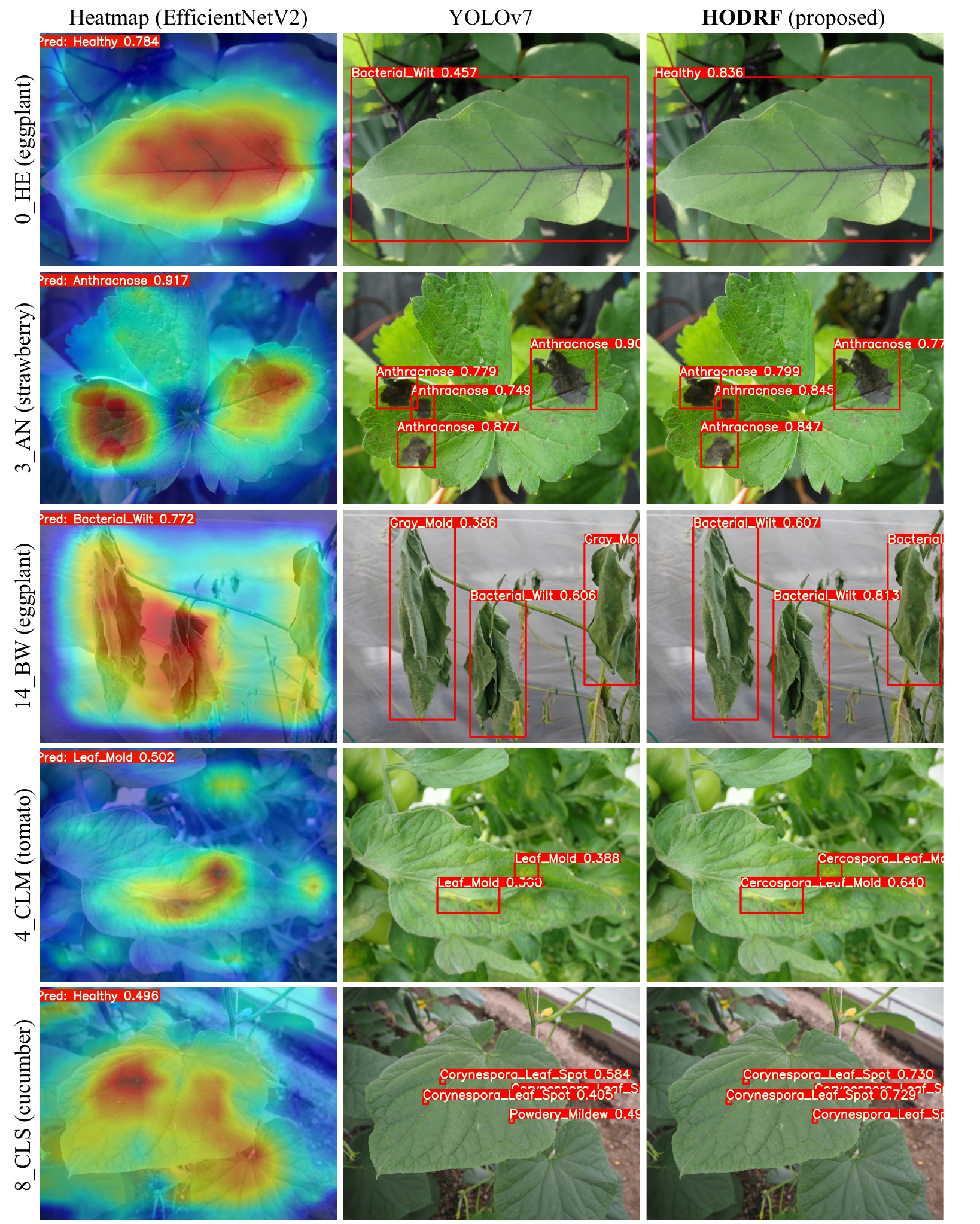}
    \caption{Example of the final diagnostic results of the three systems of the evaluation data set for a field different from the training fields.}
    \label{fig:diagcomp}
    \end{center}
\end{figure*}
Table~\ref{table2} compares the diagnostic performance, and
Fig.~\ref{fig:diagcomp} shows some diagnostic examples of this experiment. The first row shows healthy cases of eggplant, the second to fifth rows show examples of diseases, the first column shows the results of EfficientNetV2 and the heatmap as the diagnostic basis by Grad-CAM++~\cite{chattopadhay2018grad}, the second column shows the results of YOLOv7, and the third column shows example results of the proposed HODRF.
\\\indent 
Table~\ref{table2} shows that the proposed HODRF performed better diagnostically than EfficientNetV2 and YOLOv7 for all crops, whereas YOLOv7 showed a better performance in general but was inferior to EfficientNetV2 for healthy cases, as it cannot explicitly train healthy data.

\section{Discussion}
    Table~\ref{table2} shows that even with the latest CL (EfficientNetV2), diagnostic performance is still poor (macro F1 = 56.1) in crops with difficult-to-diagnose diseases such as cucumber, as in previous large-scale studies~\cite{shibuya2021validation}. One obvious reason is that CL cannot cope with diversity in the distance to the subject or in its composition, so if the training data lack sufficient diversity, the performance will be strongly influenced by how similar the test data are to the training data. YOLOv7, a state-of-the-art OD, was robust to distance variations and showed a generally strong performance, but it was inferior to EfficientNetV2 for healthy cases because it could not explicitly learn healthy data. In addition, the limited number of training data due to the high labeling cost affected the performance. The HODRF, which combines the strengths of both OD and CL methodologies, achieved the best overall performance regardless of crop type, disease, or health.
\\\indent 
In practice, the primary goal is to detect disease at an early stage. However, the ability to correctly identify a large number of health cases is also important. The HODRF achieved an F1-score of 5.8-21.5 points higher than YOLOv7 for healthy cases. As shown in the top row of Fig.~\ref{fig:diagcomp}, YOLOv7 is forced to diagnose a case as healthy when no ROI is detected, while the HODRF has the great advantage that missed healthy data can be determined by the CL that explicitly learned the healthy case at a later stage.
\\\indent 
The HODRF also had the best F1-score for disease cases across all crops (an average of 2.70 and 4.28 points better than YOLOv7 and EfficientNetV2, respectively). As shown in the second to fifth lines of Fig.~\ref{fig:diagcomp}, the HODRF assigns the area surrounding the detected ROI the diagnostic target of the CL, i.e., the model's attention can be directed to the disease signs, which is an effective means of improving diagnostic performance in cases previously difficult to diagnose.
\\\indent 
The proposed HODRF has two advantages in terms of improved robustness. First, one of the factors influencing overfitting that was mentioned in some previous studies~\cite{saikawa2019aop,wayama2023investigation} is the background, but the HODRF detects ROIs with OD and successfully suppresses these effects by diagnosing CL for the area surrounding the ROI. In addition, the CL can be trained on a wider variety of data, including healthy data, compared to OD, because the collection of training data is much easier than with OD. However, the poor diagnostic performance of the model, mainly due to intrinsic differences between training and test data (i.e., domain shift or covariate shift), is seen in difficult-to-diagnose diseases such as cucumber and should be further addressed in the future.
\subsection*{Limitations of this study}
This study did not consider multiple infections or physiological disorders, which have not been addressed by other studies.
There remains significant scope for research on enhancing diagnostic capabilities for plant diseases and insect damage, particularly in addressing the substantial gap between training data and evaluation data. Future research should focus on these areas.

\section{Conclusion}
    The proposed HODRF has achieved significant improvements in diagnostic results thanks to its effective two-stage structure, which allows it to utilize large amounts of training data, including healthy data that cannot be explicitly learned by OD, and also to adequately address the diversity in the size of the objects to be detected, which is a weakness of CL.
On the other hand, there is still room for research and development in addressing diseases with a large domain shift, where image characteristics differ fundamentally between training and test data.
Further advancements are needed to overcome problems caused by a lack of diversity in the training data.
    
\section*{Acknowledgment}
This work was supported by the Ministry of Agriculture, Forestry and Fisheries (MAFF) Japan Commissioned project study on \say{Development of pest diagnosis technology using AI} (JP17935051) and by the Cabinet Office, Public / Private R\&D Investment Strategic Expansion Program (PRISM).
\bibliographystyle{IEEEtran}
\bibliography{main}
\end{document}